\documentclass[11pt,a4paper]{article}
\usepackage[hyperref]{acl2019}
\usepackage{times}
\usepackage{latexsym}

\usepackage{url}

\usepackage[utf8]{inputenc}
\usepackage{covington}
\usepackage{graphicx}        %
                             
 \usepackage{subcaption}
 \usepackage{array}
\newcolumntype{M}[1]{>{\centering\arraybackslash}m{#1}}
\usepackage{color}
\usepackage{caption}
\usepackage{capt-of}
\usepackage{booktabs}
\usepackage{linguex}
\usepackage{amsmath}
\usepackage{ragged2e}
\usepackage{color, colortbl}

\usepackage{wrapfig}
\usepackage{supertabular}

\let\OLDthebibliography\thebibliography
\renewcommand\thebibliography[1]{
  \OLDthebibliography{#1}
  \setlength{\parskip}{2pt}
  \setlength{\itemsep}{2pt}
}

\definecolor{Gray}{gray}{0.9}
\definecolor{LightCyan}{rgb}{0.88,1,1}
\definecolor{Aquamarine}{rgb}{0.5, 1.0, 0.83}
\definecolor{Almond}{rgb}{0.94, 0.87, 0.8}
\definecolor{azure(colorwheel)}{rgb}{0.0, 0.5, 1.0}

\aclfinalcopy

\title{Meet Up! A Corpus of Joint Activity\\ Dialogues in a Visual Environment}

\author{Nikolai Ilinykh $\,$ Sina Zarrie{\ss} $\,$ David Schlangen$^\star$\\
  Dialogue Systems Group, Bielefeld University\\
  $^\star$Computational Linguistics, University of Potsdam\\
  \texttt{first.last@uni-\{bielefeld|potsdam\}.de}
}

\date{}

\begin{document}
\maketitle
\begin{abstract}
  Building computer systems that can converse about their visual environment is one of the oldest concerns of research in Artificial Intelligence and Computational Linguistics (see, for example, Winograd's 1972 SHRDLU system). Only recently, however, have methods from computer vision and natural language processing become powerful enough to make this vision seem more attainable. Pushed especially by developments in computer vision, many data sets and collection environments have recently been published that bring together verbal interaction and visual processing. Here, we argue that these datasets tend to oversimplify the dialogue part, and we propose a task---MeetUp!---that requires both visual and conversational grounding, and that makes stronger demands on representations of the discourse. MeetUp!\ is a two-player coordination game where players move in a visual environment, with the objective of finding each other. To do so, they must talk about what they see, and achieve mutual understanding. We describe a data collection and show that the resulting dialogues indeed exhibit the dialogue phenomena of interest, while also challenging the language \& vision aspect.
\end{abstract}

\section{Introduction}
\label{intro}

In recent years, there has been an explosion of interest in language \& vision in the NLP community, leading to systems and models able to ground the meaning of words and sentences in visual representations of their corresponding referents, e.g.\ work in object recognition \cite{googlenet}, image captioning \cite{fangetal:2015,devlin:imcaqui,chen2015mind,vinyals:show,Bernardietal:automatic}, referring expression resolution and generation \cite{Kazemzadeh2014,mao15,Yu2016,schlazar:acl16}, multi-modal distributional semantics \cite{kiela:2014,silberer-lapata:2014,lazaridou-pham-baroni:2015}, and many others.

While these approaches focus entirely on visual grounding in a static setup, 
a range of recent initiatives have extended exisiting data sets and models to more interactive settings.
Here, speakers do not only describe a single image or object in an isolated utterance, but engage in some type of multi-turn interaction to solve a given task \cite{das2017visual,vries2017guesswhat}. 
In theory, these data sets should allow for more dynamic approaches to \textit{grounding} in natural language interaction, where words or phrases do not simply have a static multi-modal meaning (as in existing models for distributional semantics, for instance),  but, instead, where the meaning of an utterance is \emph{negotiated} and \emph{established} during interaction. Thus, ideally, these data sets should lead to models that combine visual grounding in the sense of \newcite{harnard:grounding} and conversational grounding in the sense of \newcite{clark1991grounding}.

In practice, however, it turns out to be surprisingly difficult to come up with data collection set\-ups that lead to interesting studies of both these aspects of grounding.
 Existing tasks still adopt a very rigid interaction protocol, where e.g.\ an asymmetric interaction between a question asker and a question answerer produces uniform sequences of question-answer pairs (as in the ``Visual Dialogue'' setting of \newcite{das2017visual} for instance). 
 Here, it is impossible to model e.g.\ turn-taking, clarification, collaborative utterance construction, which are typical phenomena of conversational grounding in interaction \cite{clark1996using}.
Others  tasks follow the traditional idea of the \emph{re\-ference game} \cite{Rosenberg1964,clarkwilkes:ref} in some way, but try to set up the game such that the referent can only be established in a sequence of turns \cite[e.g.][]{vries2017guesswhat}. 
While this approach leads to goal-oriented dialogue,
the goal  is still directly related to reference and visual grounding.
However, realistic, every-day communication between human speakers rarely centers entirely around establishing reference. It has been argued in the literature that reference production radically changes if it is the primary goal of an interactive game, rather than embedded in a dialogue that tries to achieve a more high-level communicative goal \cite{stent2011computational}.

Another strand of recent work extends the environments about which the language can talk to (simulated) 3D environments (\citet{habitat19arxiv,savva2017minos}; see \citet{byron2007generating} for an early precursor). On the language side, however, the tasks that have been proposed in these environments allow only limited interactivity (navigation, e.g. \citet{Anderson2018,Ma2019}; question answering, \citet{dasetal:eqa}).

\begin{figure}[h]
\centering
   \fbox{\includegraphics[scale=.34]{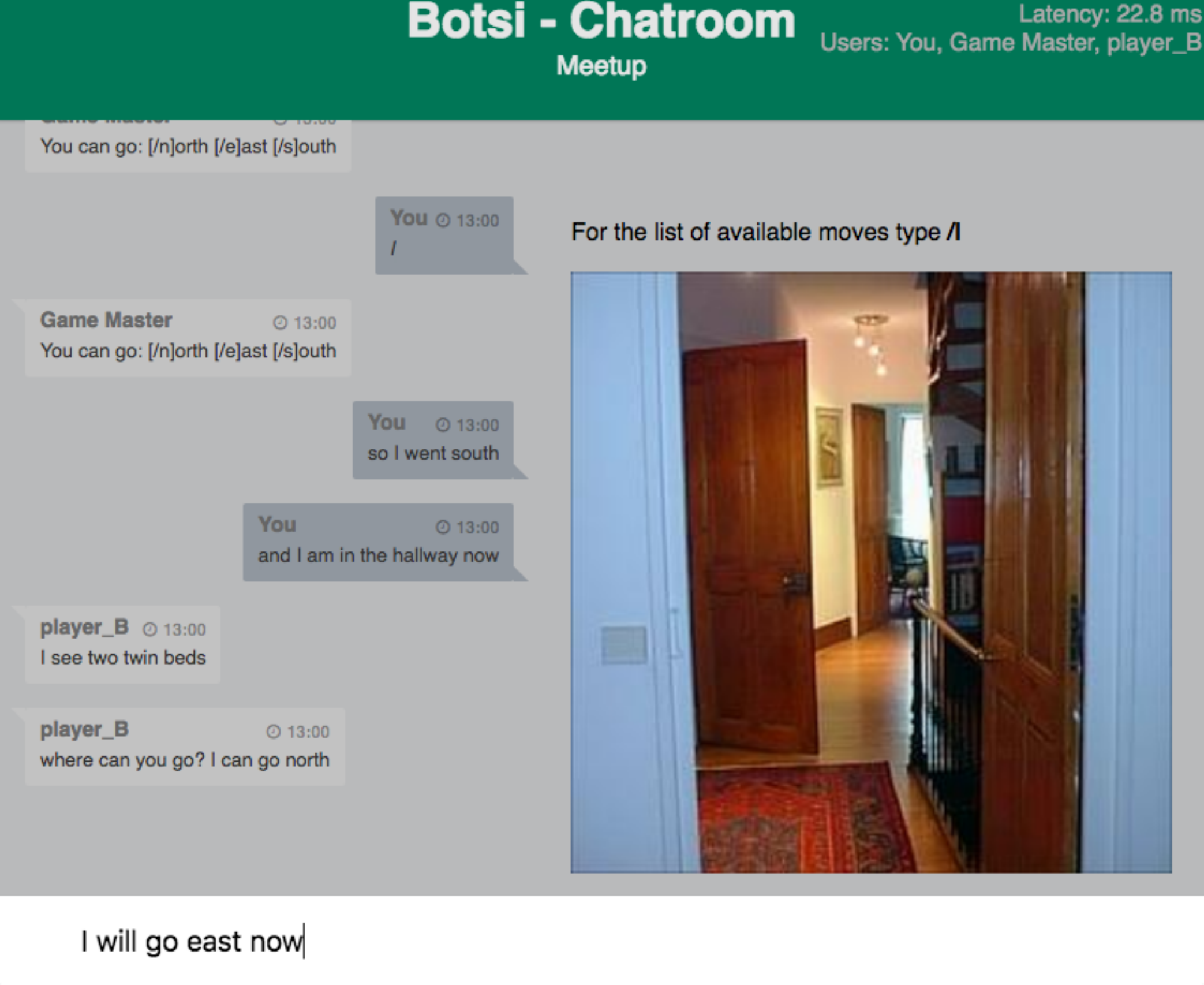}}
   \caption{The game interface}
\label{fig:meetup}
\end{figure}

What is lacking in these tasks is a real sense of the interaction being a \emph{joint task} for which both participants are equally responsible, and, phrased more technically, any need for the participants to jointly attempt to track the dialogue state.
In this paper, we propose a new task, MeetUp!, for visually grounded interaction, which is aimed at collecting conversations about and within a visual world, in a collaborative setting.
(Figure~\ref{fig:meetup} gives a view of the game interface and an excerpt of an ongoing interaction.)

Our setup extends recent efforts along three main dimensions: 1) the task's main goal can be defined independently of reference, in high-level communicative terms (namely ``try to meet up in an unknown environment''), 2) the task is symmetric and does not need a rigid interaction protocol (there is no instruction giver/follower), 3) the requirement to \emph{agree} on the game state (see below) ensures that the task is a true \emph{joint activity} \cite{clark:ul}, which in turn brings out opportunity for \emph{meta-semantic} interaction and negotiation about perceptual classifications (``there is a mirror'' --  ``hm, could it be a picture?''. This is an important phenomenon absent from all major current language \& vision datasets.

This brings our dataset closer to those of unrestricted natural situated dialogue, e.g.\ \cite{anderson1991hcrc,fernangen:sigd07,tokunaga2012rex,zarriess2016pentoref}, while still affording us some control over the expected range of phenomena, following our design goal of creating a challenging, but not too challenging modelling resource. The crowd-sourced nature of the collection also allows us to create a resource that is an order of magnitude larger than those just mentioned.\footnote{%
  \citet{Haber2019} present a concurrently collected dataset that followed very similar aims (and is even larger); their setting however does not include any navigational aspects and concentrates on reaching agreement of whether images are shared between the participants or not. 
}

We present our data collection of over 400 dialogues in this domain, providing an overview of the characteristics and an analysis of some occuring phenomena.
Results indicate that the task leads to rich, natural and varied dialogue where speakers use a range of strategies to achieve communicative grounding.
The data is available from \url{https://github.com/clp-research/meetup} .

\section{The Meet Up Game}
\label{sec:mu}

\begin{figure}[h]
  \includegraphics[scale=.4]{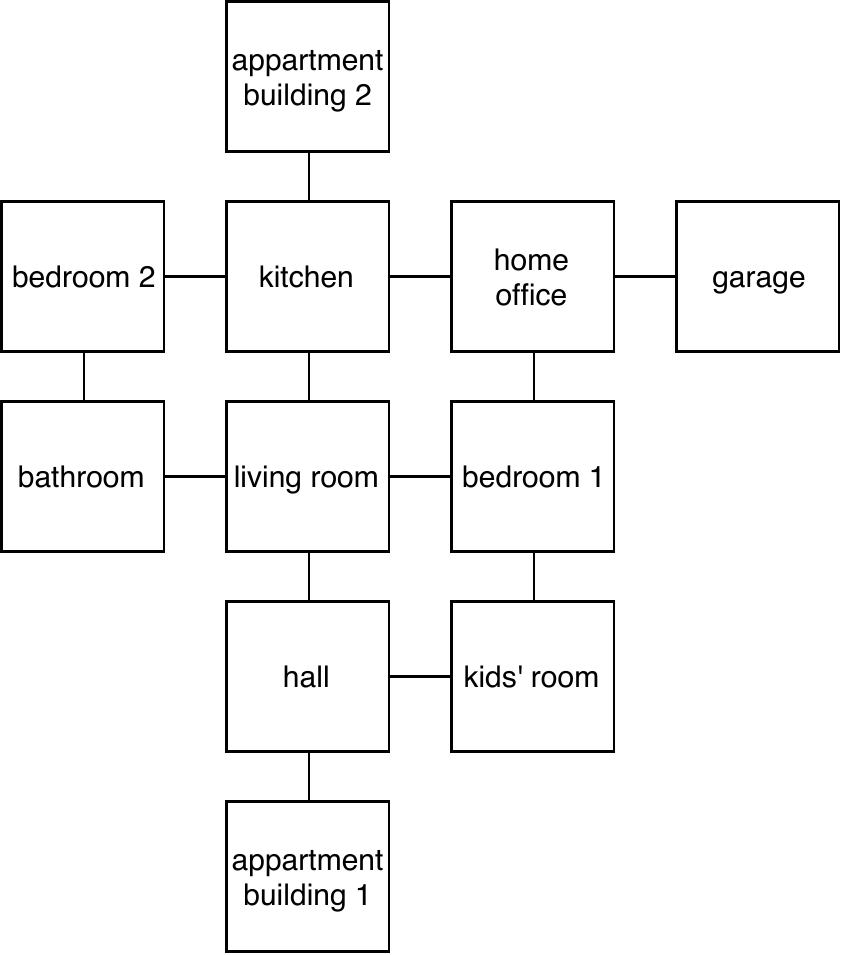}
  $\;$
  \includegraphics[scale=.4]{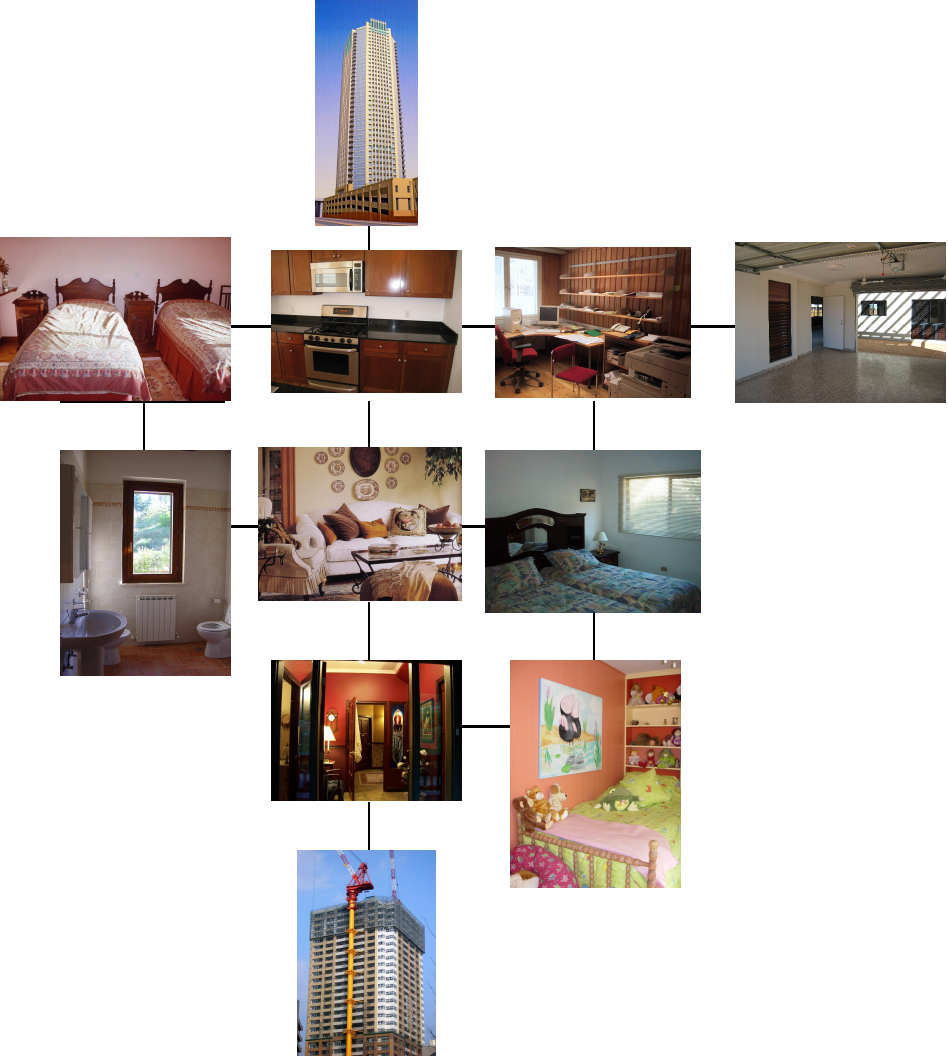}
  \caption{An abstract layout with room types (left), and a full gameboard with assigned images (right).}       
\label{fig:gb}
\end{figure}

MeetUp!\ is a two-player coordination game. In the discrete version described here, it is played on a gameboard that can be formalised as a connected subgraph of a two-dimensional grid graph.\footnote{%
  The game could also be realised in an environment that allows for continuous movement and possibly interaction with objects, for example as provided by the simulators discussed above. This would complicate the navigation and visual grounding aspects (bringing those more in line with the ``vision-and-language navigation task''; \cite[e.g.][]{Anderson2018,Ma2019}), but not the coordination aspect. As our focus for now is on the latter, we begin with the discrete variant.
}
See Figure~\ref{fig:gb} for an example.

Players are located at vertices in the graph, which we call ``rooms''.
Players never see a representation of the whole gameboard, they only see their current room (as an image). They also do not see each other's location.
The images representing rooms are of different types; here, different types of real-world scenes, such as ``bathroom'', ``garage'', etc., taken from the ADE20k corpus collected by \newcite{zhou2017scene}.
Players can move from room to room, if there is a connecting edge on the gameboard. On entering a room, the player is (privately) informed about the available exit directions as cardinal directions, e.g. ``north'', ``south'', etc., and (privately) shown the image that represents the room.
Players move by issuing commands to the game; these are not shown to the other player.

The goal of the players is to be in the same location, which means they also have to be aware of that fact. In the variant explored here, the goal is constrained in advance in that the meetup room has to be of a certain type previously announced to the players; e.g., a kitchen.
The players can communicate via text messages. As they do not see each other's location, they have to describe the images they see to ascertain whether or not they are currently in the same room, and move to a different room if they decide that they aren't. If they have reached the conclusion that they are, they can decide to end the game, which they do via a special command. If they are then indeed in the the same room, and it is of the target type, the game is counted as a success, of which they are informed.
The gameboard can be arranged such that there is type-level ambiguity; for example, there may be more than one room of type ``bedroom'' (as in Figure~\ref{fig:gb}).

The game as implemented does not impose strict turn taking on the players; however, messages are only shown to the other player once they are sent via pressing the return key, as is usual in chat tools. There is thus no possibility for perceivably overlapping actions, but it may happen that both players have been typing at the same time and the message that is received second is not a response to the first.

To make this more concrete, and to explain our expectations with respect to phenomena and required capabilities, we show a realistic, but compressed and constructed example of an interaction in this domain in the following. We will discuss attested examples from our data collection further below.

\ex. \label{ex:constructed}
\small
\a. \label{ex:c-gm}
Game Master: You have to meet in a room of type \emph{utility room}.
\b. \label{ex:c-aini}
A: Hi. I'm in a bedroom with pink walls.
\b. \label{ex:c-bini}
B: I seem to be in a kitchen.
\b. \label{ex:c-strat}
A: I'll go look for a utility room.
\b. \label{ex:c-nav-1}
A (privately): \emph{north}
\b. A (privately): \emph{west}
\b. \label{ex:c-nav-3}
B (privately): \emph{east}
\b. \label{ex:c-cclar-1}
A: Found a room with a washing machine. Is that a utility room?
\b. \label{ex:c-cclar-2}
B: Was wondering as well. Probably that's what it is.
\b. B: \label{ex:c-strat-b}
I'm in the pink bedroom now. I'll come to you.
\b. B (privately): \emph{north}
\b. B (privately): \emph{west}
\b. B: \label{ex:c-posmir-1}
Poster above washing machine?
\b. A: \label{ex:c-posmir-2}
Mine has a mirror on the wall.
\b. B: \label{ex:c-posmir-3}
yeah, could be mirror. Plastic chair?
\b. A: \label{ex:c-basket}
And laundry basket.
\b. A: \emph{done}
\b. B: Same
\b. B: \emph{done}

In \ref{ex:c-gm}, the Game Master (realised as a software bot in the chat software) gives the type constraint for the meetup room, which sets up a \textbf{classification task} for the players, namely to identify rooms of this type.
\ref{ex:c-aini} and \ref{ex:c-bini} illustrate a common strategy (as we will see below), which is to start the interaction by providing state information that potentially synchronises the mutual representations. This is done through the production of \textbf{high-level descriptions of the current room}; for which the agents must be capable of providing \emph{scene categorisations}.
\ref{ex:c-strat} and \ref{ex:c-strat-b} show, among other things, the \textbf{coordination of strategy}, by announcing plans for action.
In \ref{ex:c-nav-1} -- \ref{ex:c-nav-3}, private navigation actions are performed, which here are both \textbf{epistemic actions} (changing the environment to change perceptual state) as well as \textbf{pragmatic actions} (task level actions that potentially advance towards the goal), in the sense of \citet{Kirsh1994}. \ref{ex:c-cclar-1} and \ref{ex:c-cclar-2}, where the classification decision itself and its basis is discussed (``what is a utility room?''); and \ref{ex:c-posmir-1}--\ref{ex:c-posmir-3}, where a classification decision is revised (\emph{poster} to \emph{mirror}), illustrate the potential for \textbf{meta-semantic interaction}.
This is an important type of dialogue move \cite{schlangen:justification}, which is entirely absent from most other language and vision datasets and hence outside of the scope of models trained on them.
\ref{ex:c-strat-b}, also illustrates the need for \textbf{discourse memory}, through the co-reference to the earlier mentioned room where A was at the start.
Finally, \ref{ex:c-basket} as reply to \ref{ex:c-posmir-3} shows how in conversational language, \textbf{dialogue acts} can be \textbf{performed indirectly}.

As we have illustrated with this constructed example, the expectation is that this domain challenges a wide range of capabilities; capabilities which so far have been captured separately (e.g., visual question answering, scene categorisation, navigation based on natural language commands, discourse co-reference), or not at all (discussion and revision of categorisation decisions). We will see in the next section whether this is borne out by the data.

\section{Data Collection}

To test our assumptions, and to later derive models for these phenomena, we collected a larger number of dialogues in this domain (430, to be precise). We realised the MeetUp game within the \emph{slurk} chat-tool \cite{slurk.semdial18}, deployed via the Amazon Mechanical Turk platform.

We constructed maps for the game in three steps. First, we create a \emph{graph} through a random walk over a grid graph, constrained to creating 10 nodes. The nodes are then assigned room types, to form what we call a \emph{layout}.
 We identified 48 categories from the ADE20k corpus that we deemed plausible to appear in a residential house setting, from which we designated 20 categories as possible (easy to name) target types and the remaining 28 as distractor types.
 Additionally, we identified 24 plausible outdoor scene types, from which we sampled for the leaf nodes. The full set is given in the Appendix.
 We designate one type per layout to be the target type; this type will be assigned to 4 nodes in the graph, to achieve type ambiguity and potentially trigger clarification phases. We then sample actual images from the appropriate ADE20k categories, to create the \emph{gameboards}. In a final step, we randomly draw separate starting positions for the players, such that both of the players start in rooms not of the target type. For each run of the game, we randomly create a new gameboard following this recipe.

We deployed the game as a web application, enlisting workers via the Mechanical Turk platform. After reading a short description of the game (similar to that at the beginning of Section~\ref{sec:mu}, but explaining the interface in more detail), workers who accepted the task were transferred to a waiting area in our chat tool. If no other worker appeared within a set amount of time, they were dismissed (and payed for their waiting time). Otherwise, the pair of users was moved to another room in the chat tool and the game begun. Player were payed an amount of \$0.15 per minute (for a maximum of 5 minutes per game), with a bonus of \$0.10 for successfully finishing the game (as was explained from the start in the instruction, to provide an additional incentive).\footnote{%
  By the time of the conference, we will publish the code required to run this environment, as well as the data that we collected.
}

\section{Results}

\subsection{Descriptive Statistics}
\label{sec:desc}

Over a period of 4 weeks, we collected 547 plays of the game. Of these, 117 (21\%) had to be discarded because one player left prematurely or technical problems occurred, which left us with 430 completed dialogues. Of these, 87\% ended successfully (players indeed ending up in the same room, of the correct type), 10\% ended with the players being in different rooms of the correct type; the remaining 3\% ended with at least one player not even being in a room of the target type. Overall, we spent around \$700 on the data collection.

\begin{figure}[h]
\centering
   \includegraphics[scale=.5]{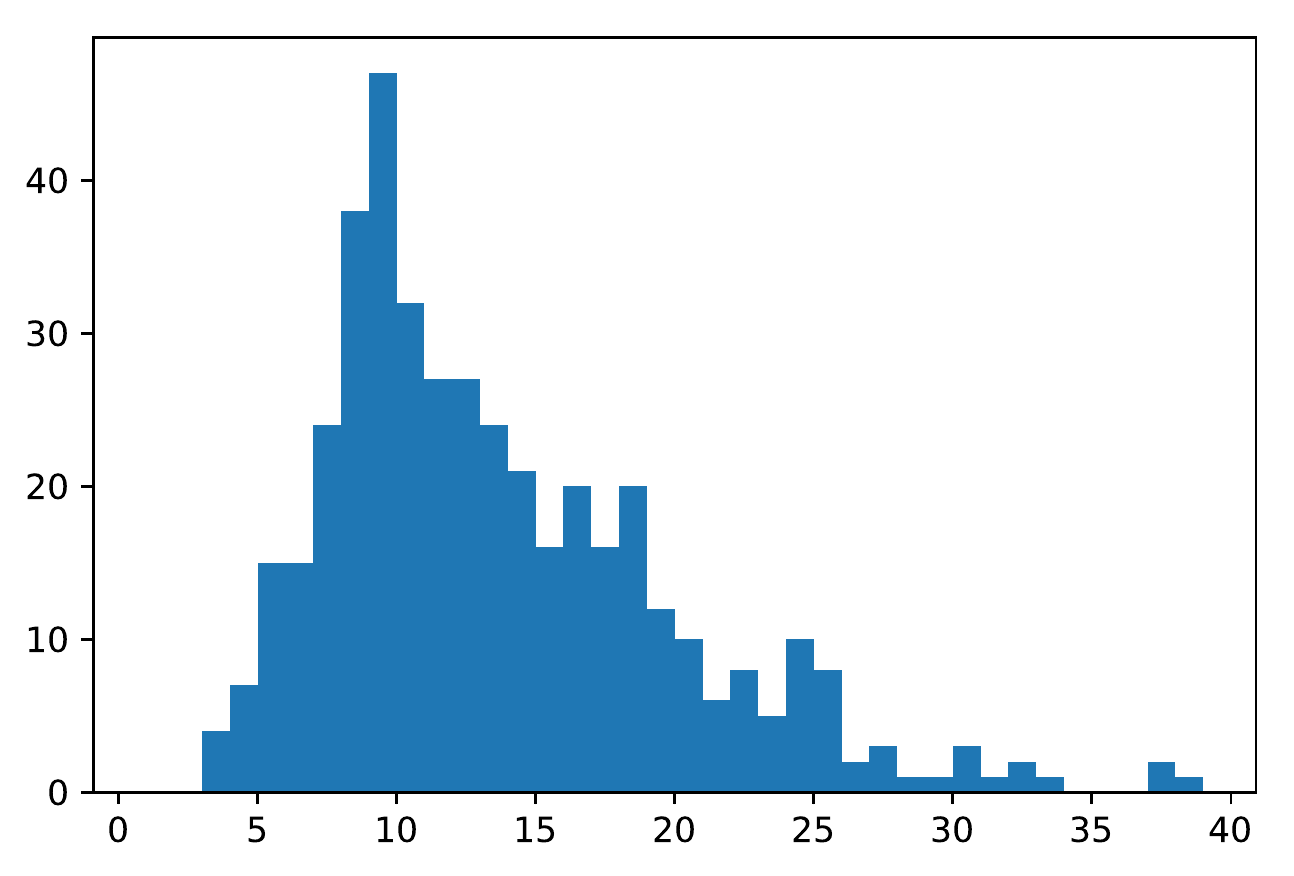}
   \vspace*{-.5\baselineskip}
   \caption{Histogram of number of turns per dialogue}
\label{fig:dlhist}
\end{figure}

The average length of a dialogue was 13.2 turns (66.9 tokens), taking 165 seconds to produce. (The distribution of lengths is shown in Figure~\ref{fig:dlhist}.) Altogether, we collected 5,695 turns, of an average length of 5.1 tokens. Over all dialogues, 2,983 word form types were introduced, leading to a type/token ratio of 0.10. The overlap of the vocabularies of the two players (intersection over union) ranged from none to 0.5, with a mean of 0.11.

On average, in each dialogue 28.3 navigation actions were performed. (Resulting in  a \textsc{move}/\textsc{say} ratio of a little over 2 to 1). The median time spent in a room was 12.2 secs. On average, each player visited 5.9 rooms without saying anything; when a player said something while in a room, they produced on average 3.5 turns. It hence seems that, as expected,  players moved through some rooms without commenting on them, while spending more time in others.

We calculated the contribution ratio between the more talkative player and the less talkative one in each dialogue, which came out as 2.4 in terms of tokens, and 1.7 in terms of turns. This indicates that there was a tendency for one of the players to take a more active role. To provide a comparison, we calculated the same for the (role-asymmetric) MapTask dialogues \cite{anderson1991hcrc},\footnote{%
  Using the transcripts provided at \url{http://groups.inf.ed.ac.uk/maptask/maptasknxt.html}.
}
finding a 2.8 token ratio and a 1.3 turn ratio.

\begin{figure}[h]
\centering
   \includegraphics[scale=.5]{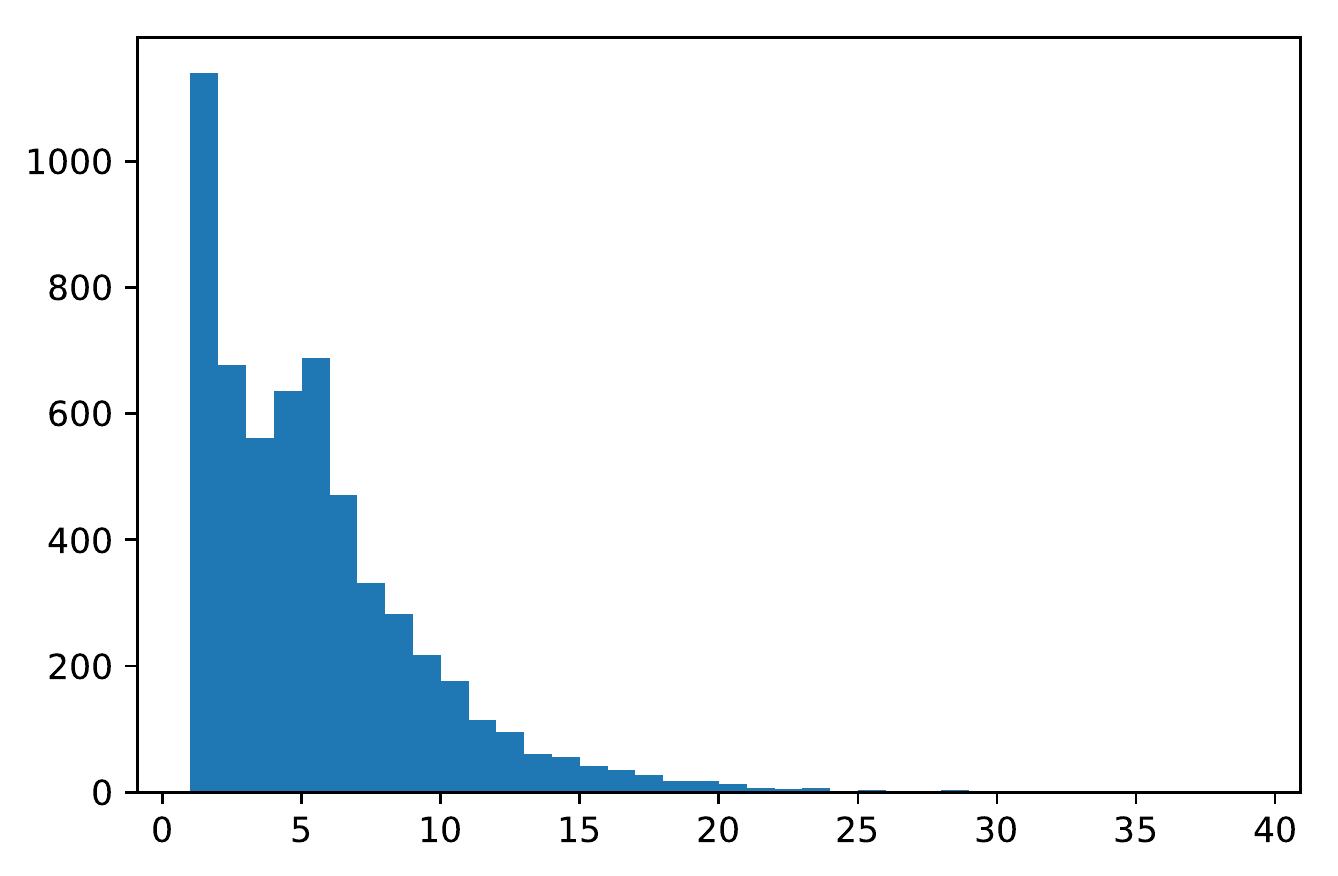}
   \vspace*{-.8\baselineskip}
   \caption{Histogram of number of tokens per turn}
   \vspace*{-1.5\baselineskip}
\label{fig:hist}
\end{figure}

\begin{figure}[h]
\centering
   \includegraphics[scale=.5]{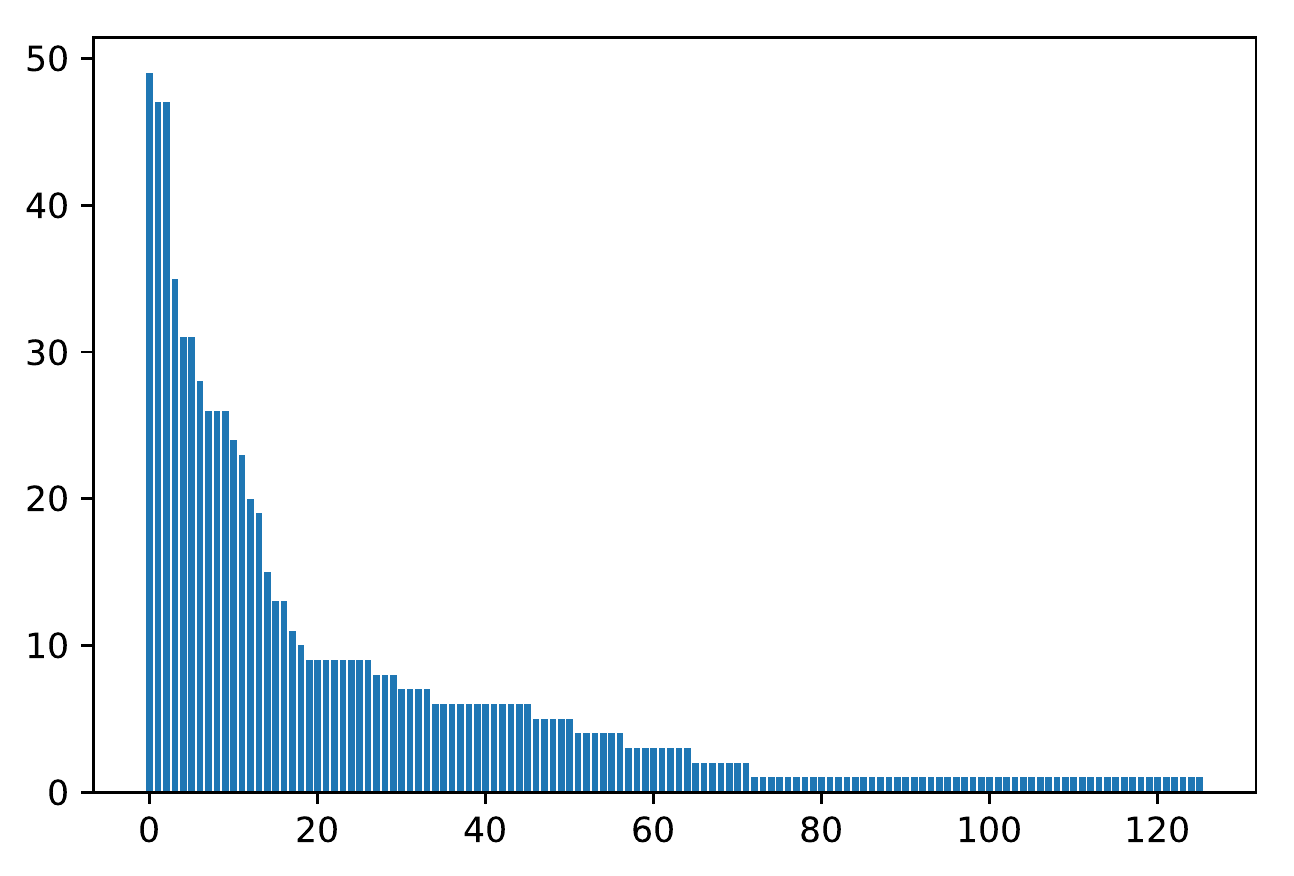}
   \vspace*{-.8\baselineskip}
   \caption{Number of Games Played, by Worker}
   \vspace*{-.7\baselineskip}
\label{fig:repeat}
\end{figure}

Crosstalk occurs: On average, there are 1.4 instances of one turn coming within two seconds or less than the previous one (which we arbitrarily set as the threshold for when a turn is likely not to be a reaction to the previous one, but rather has been concurrently prepared). The mean pause duration between turns of different speakers is 11.2 secs -- with a high standard deviation of 9.46, however. This is due to the structure of the dialogues with phases of intense communicative activity, when a matching decision is made, and phases of individual silent navigation. If we only take transition times within the first 3 quartiles, the average transition time is 5.04 secs.

As Figure~\ref{fig:hist} indicates, most turns are rather short, but there is a substantial amount of turns that contain 4 or more tokens.

Figure~\ref{fig:repeat} shows a frequency distribution of number of games played, by crowdworker. Overall, we had 126 distinct participants (as indicated by AMT ID). Our most prolific worker participated in 49 games, and the majority of workers played in more than one game. In only 22 games, two novices played with each other. In 81 games, there was one novice, and in 305 games, both players had played before. (For a few games we could not reconstruct the workerIDs for technical reasons, so this does not sum up to 430.)

\subsection{Examples}
\label{sec:exdials}

\begin{figure*}[h!]
  {\small
    \includegraphics[scale=.6]{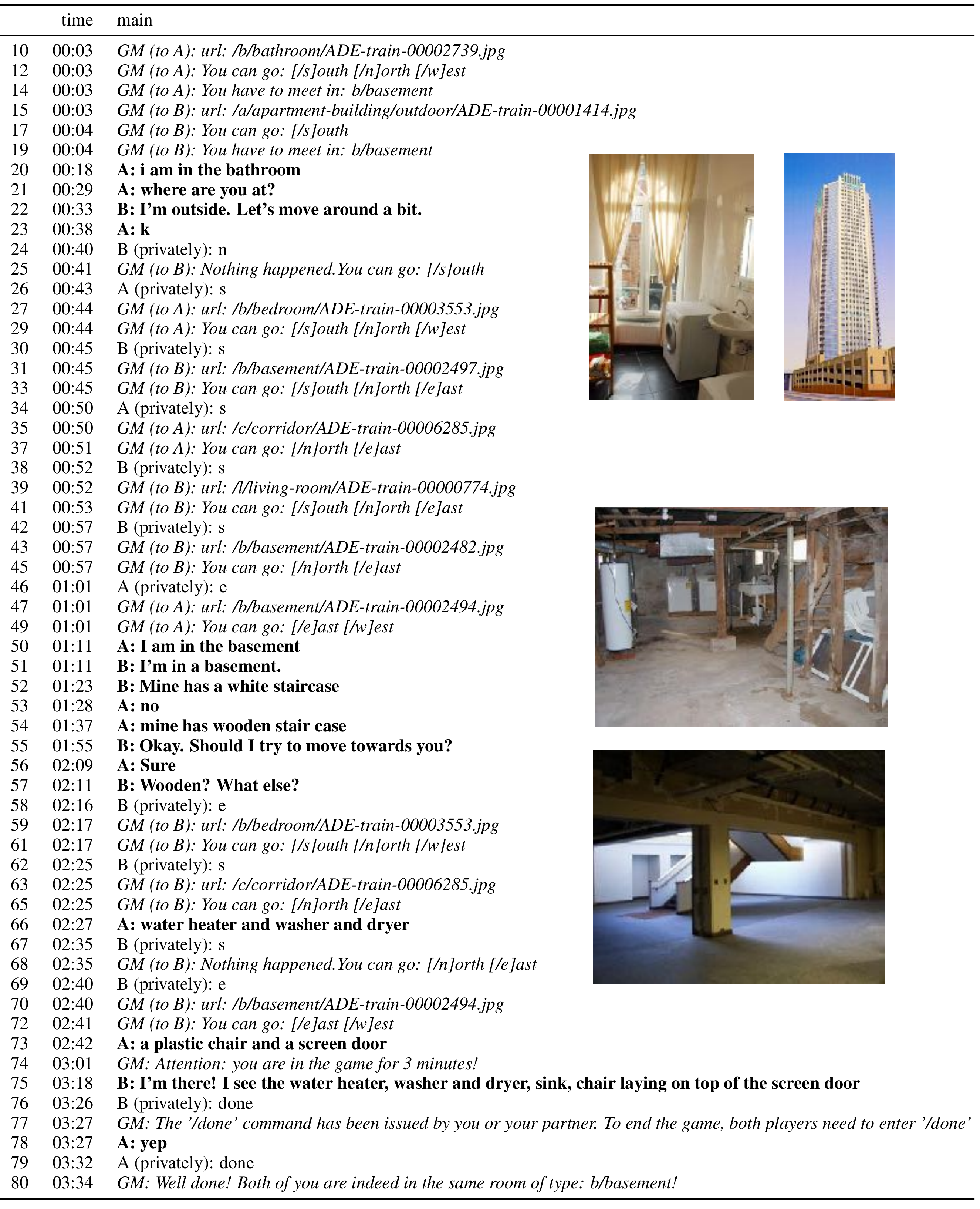}
  }
  \vspace*{-1\baselineskip}
  \caption{One Example Dialogue (\texttt{mux36}), with Images Overlayed}
  \label{tab:staircase}
\end{figure*}

Figure~\ref{tab:staircase} shows a full interaction from the corpus. The public actions are represented in bold font, private actions are marked with ``privately'', and responses by the Game Master are shown in italics. This example has many of the features illustrated with the constructed example \ref{ex:constructed} shown earlier. In lines 20 and 22, the players begin the game by providing high-level categorisations of their current positions, in effect synchronising their mutual game representations. Lines 22 and 23 then show coordination of game playing strategy. After separately moving around, individually solving the categorisation task--by moving through rooms that are not of the right type---the players resume interaction in lines 50ff.\ (with 50/51 showing crosstalk). Line 54 provides a justification for the negative conclusion from line 53, by providing information that contradicts l.\ 52. After more coordination of strategy in l.s 55 \& 56, player B explicitly queries for more information. In line 75, player A justifies their positive verdict by confirming details previously given by B, extending it with even more details. B confirms explicitly in 78, before also chosing \textsc{solve}.

The excerpt from another dialogue in \ref{ex:uncertainty} shows an example of classification uncertainty being negotiated and dealt with.

\ex. \label{ex:uncertainty}
(Excerpt from \texttt{mux39})\\
A: i think i am in a basement \\
B: i think i might be too. \\
A: maybe not though \\
A: wood panel? \\
A: two doors? \\
B: there's a tan couch, and a tan loveseat/chair. brown coffee table. bar. tv \\
B: nope, different room \\
A: ok i am not there \\
B: want me to meet you, or do you want to meet me? \\
A: i think mine is more basement like \\
B: okay. i'll try to find it. \\

\subsection{Phases and Phenomena}
\label{sec:phases}

\begin{figure}[h]
\centering
   \includegraphics[scale=.5]{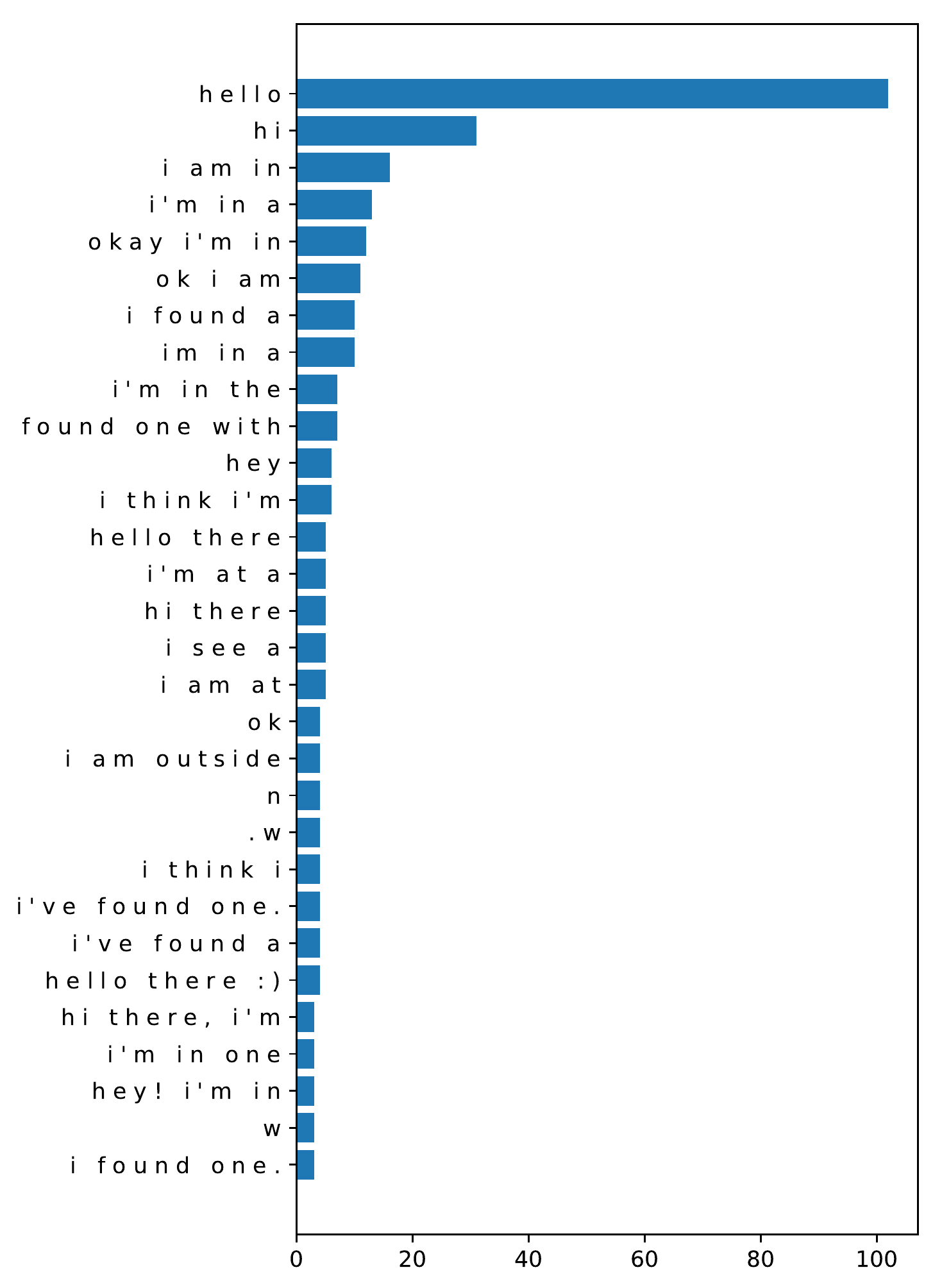}
   \vspace*{-.5\baselineskip}
   \caption{Prefixes of first turns}
   \vspace*{-.5\baselineskip}
\label{fig:firstt}
\end{figure}

\begin{figure}[h]
\centering
   \includegraphics[scale=.5]{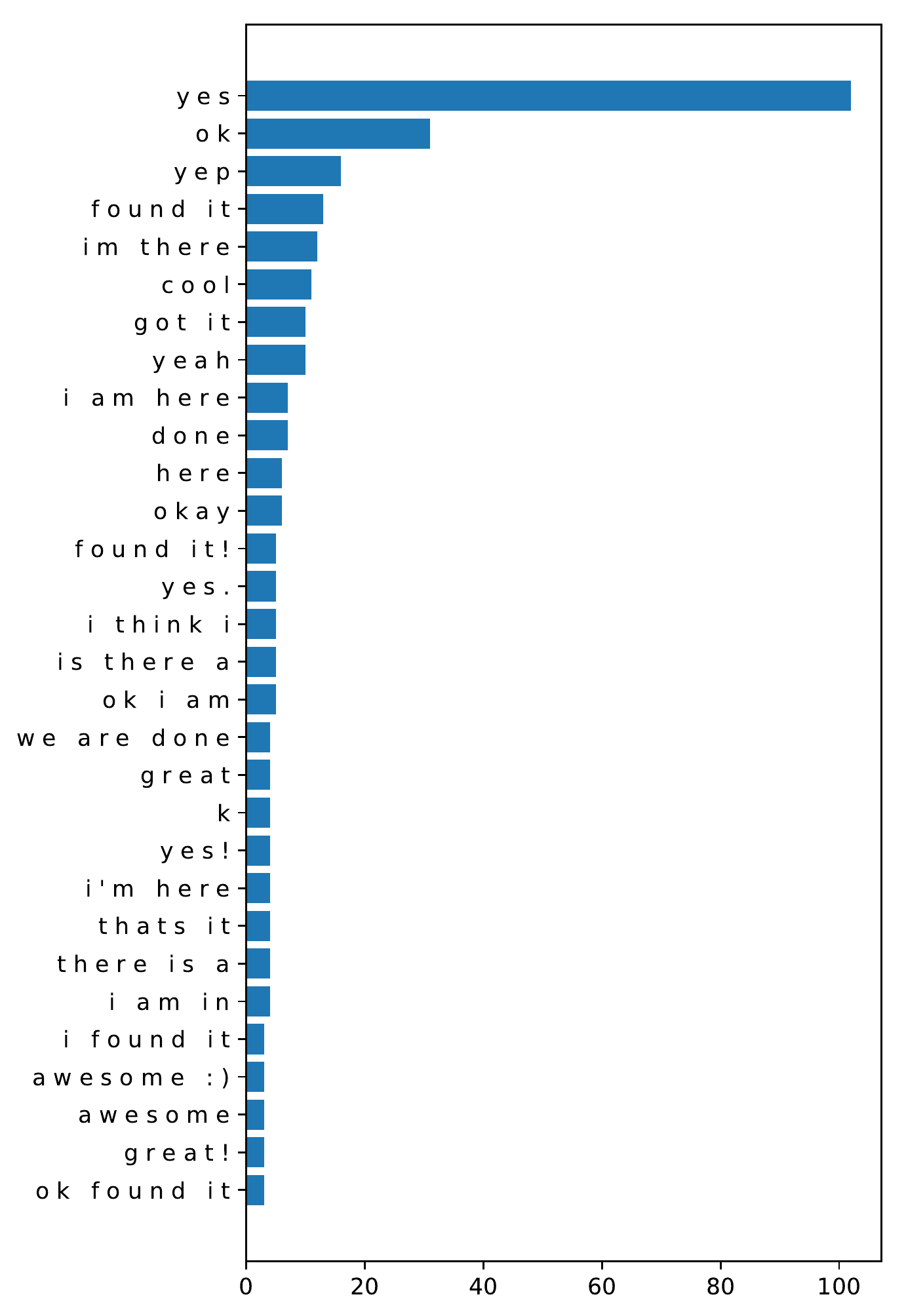}
   \vspace*{-.5\baselineskip}
   \caption{Prefixes of final turns (before \emph{done})}
   \vspace*{-2\baselineskip}
\label{fig:done}
\end{figure}

\noindent
Figure~\ref{fig:firstt} shows the most frequent beginnings of the very first turn in each dialogue. As this indicates, when not opening with a greeting, players naturally start by locating themselves (as in the example we showed in full). Figure~\ref{fig:done} gives a similar view of the final turn, before the first \emph{done} was issued. This shows that the game typically ends with an explicit mutual confirmation that the goal condition was reached, before this was indicated to the game.

\begin{figure}[h]
\centering
   \includegraphics[scale=.5]{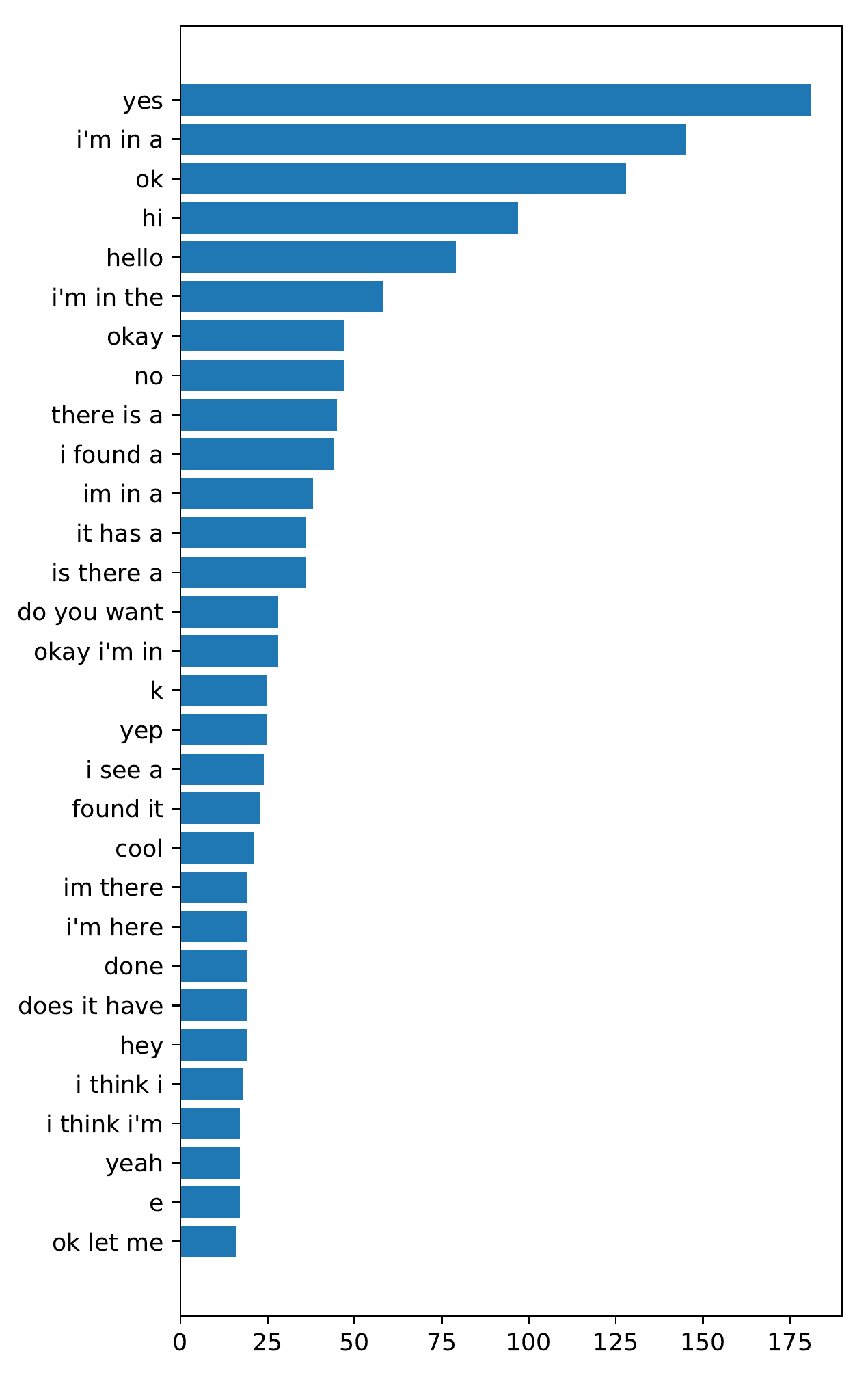}
   \vspace*{-.5\baselineskip}
   \caption{Most frequent turn beginnings}
   \vspace*{-2\baselineskip}
\label{fig:pref}
\end{figure}

\begin{figure}[h]
\centering
   \hspace*{-.9\baselineskip}
   \includegraphics[scale=.5]{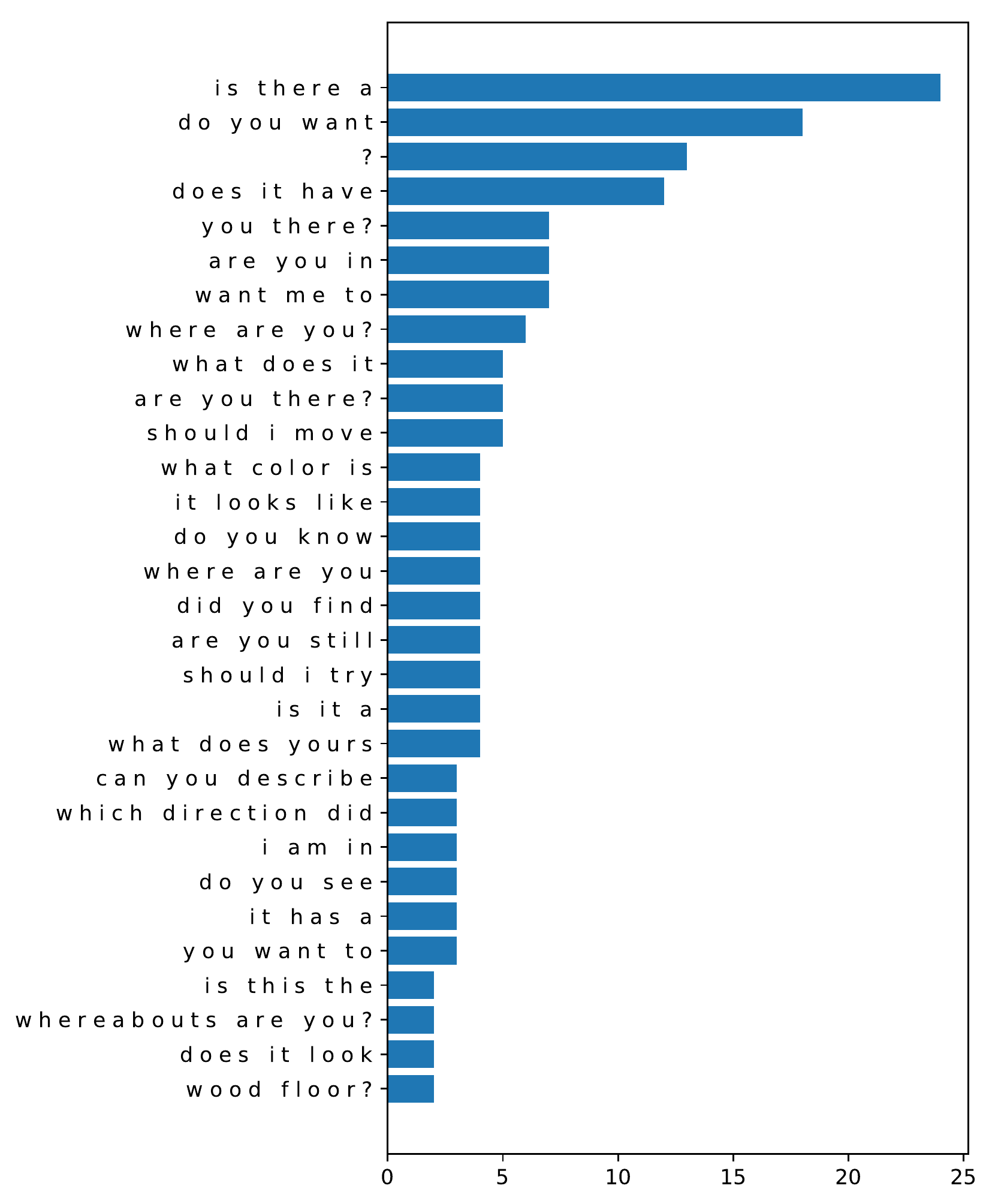}
   \vspace*{-1.5\baselineskip}
   \caption{Prefixes of questions (utt.s ending in ``?'')}
   \vspace*{-2\baselineskip}
\label{fig:questions}
\end{figure}

What happens inbetween? Figure~\ref{fig:pref} shows the most frequent overall turn beginnings. As this illustrates, besides the frequent positive replies (``yes'', ``ok''; indicating a substantial involvement of VQA-like interactions), the most frequent constructions seem to locate the speaker (``I'm in a'') or talk about objects (``I found a'', ``there is a'', ``is there a''). Using the presence of a question mark at the end of the turn as a very rough proxy, we find 615 questions over all dialogues, which works out as 1.43 on average per dialogue. Taking only the successfull dialogues into account, the number is slightly higher, at 1.48. Figure~\ref{fig:questions} shows the beginnings of these turns.

\section{Modelling the Game}
\label{sec:mod}

The main task of an agent playing this game can be modelled in the usual way of modelling agents in dynamic environments \cite{suttonbarto:rl}, that is, as computing the best possible next action, given what has been experienced so far. The questions then are what the range of possible actions is, what the agent needs to remember about its experience, and what the criteria might be for selecting the best action.

In the action space, the clearest division is between actions that are directly observable by the other player---actions of type \textsc{say}---and actions that are targeted at changing the observable game state for the agent itself: actions of type \textsc{move} and the \textsc{end} action. Since we did not restrict what the players could say, there is an infinite number of \textsc{say} actions (see \citet{Cote2018} for a formalisation of such an action space).

The total game state consists of the positions of the players on the gameboard. Of this, however, only a part is directly accessible for either agent, which is their own current position. The topology of the network must be remembered from experience, if deemed to be relevant. (From observing the actions of the players in the recorded dialogues, it seems unlikely that they attempted to learn the map; they are however able to purposefully return to earlier visited rooms.) More importantly, the current position of the other player is only indirectly observable, through what they report about it. Finally, as we have seen in the examples above, the players often negotiate and agree on a current strategy (e.g., ``I find you'', ``you find me'', ``we walk around''). As this guides mutual expectations of the players, this is also something that needs to be tracked. On the representation side, we can then assume that an agent will need to track a) their own history of walking through the map (raising interesting questions of how detailed such a representation needs to be or should be made; an artificial agent could help itself by storing the full image for later reference, which would presumably be not enitirely plausible cognitively); b) what has been publicly said and hence could be antecedent to later co-references; c) what they infer about the other player's position; and d) what they assume the current agreed upon strategy is. This clearly is a challenging task; we will in future work first explore hybrid approaches that combine techniques from task-oriented dialogue modelling \cite{williamsyoung:POMDPs,bussetal:subutt} with more recent end-to-end approaches \cite{Cote2018,Urbanek2019}.

\section{Conclusions}
\label{sec:conc}

We have presented a novel situated dialogue task that brings together visual grounding (talking about objects in a scene), conversational grounding (reaching common ground), and discourse representation (talking about objects that were introduced into the discourse, but aren't currently visible). An agent mastering this task will thus have to combine dialogue processing skills as well as language and vision skills. We hence hope that this task will lead to the further development of techniques that combine both. Our next step is to scale up the collection, to a size where modern machine learning methods can be brought to the task. Besides use in modelling, however, we also think that the corpus can be a valuable resource for linguistic investigations into the phenomenon of negotiating situational grounding.

\appendix

\section{Room Types}
\label{sec:rtypes}
\scriptsize
\hspace*{-4ex}
\begin{tabular}{|m{\columnwidth}|}
	\hline
\textbf{1. Target room types:} bathroom, bedroom, kitchen, basement, nursery, attic, childs\_room,
playroom, dining\_room, home\_office, staircase, utility\_room, living\_room,
jacuzzi/indoor, doorway/indoor, locker\_room, wine\_cellar/bottle\_storage,
reading\_room, waiting\_room, balcony/interior
\\
\hline
\textbf{2. Distractor room types:} home\_theater, storage\_room, hotel\_room, music\_studio, computer\_room,
street, yard, tearoom, art\_studio, kindergarden\_classroom, sewing\_room, shower,
veranda, breakroom, patio, garage/indoor, restroom/indoor, workroom, corridor, game\_room,
poolroom/home, cloakroom/room, closet, parlor, hallway, reception, carport/indoor, hunting\_lodge/indoor
    \\
    \hline
\textbf{3. Outdoor room types (nodes with a single entry point):} garage/outdoor, apartment\_building/outdoor, jacuzzi/outdoor, doorway/outdoor, restroom/outdoor,
swimming\_pool/outdoor, casino/outdoor, kiosk/outdoor, apse/outdoor, carport/outdoor, flea\_market/outdoor, chicken\_farm/outdoor, washhouse/outdoor, cloister/outdoor, diner/outdoor, kennel/outdoor, hunting\_lodge/outdoor, cathedral/outdoor, newsstand/outdoor, parking\_garage/outdoor, convenience\_store/outdoor, bistro/outdoor, inn/outdoor, library/outdoor
\\
\hline
  \end{tabular}
\label{tab:types}

\bibliographystyle{acl_natbib}
{\footnotesize
\bibliography{meetup}

\begin{thebibliography}{42}
\expandafter\ifx\csname natexlab\endcsname\relax\def\natexlab#1{#1}\fi

\bibitem[{Anderson et~al.(1991)Anderson, Bader, Bard, Boyle, Doherty, Garrod,
  Isard, Kowtko, McAllister, Miller et~al.}]{anderson1991hcrc}
Anne~H Anderson, Miles Bader, Ellen~Gurman Bard, Elizabeth Boyle, Gwyneth
  Doherty, Simon Garrod, Stephen Isard, Jacqueline Kowtko, Jan McAllister, Jim
  Miller, et~al. 1991.
\newblock The hcrc map task corpus.
\newblock \emph{Language and speech}, 34(4):351--366.

\bibitem[{Anderson et~al.(2018)Anderson, Wu, Teney, Bruce, Johnson,
  S{\"{u}}nderhauf, Reid, Gould, and van~den Hengel}]{Anderson2018}
Peter Anderson, Qi~Wu, Damien Teney, Jake Bruce, Mark Johnson, Niko
  S{\"{u}}nderhauf, Ian Reid, Stephen Gould, and Anton van~den Hengel. 2018.
\newblock \href {http://arxiv.org/abs/1711.07280} {{Vision-and-Language
  Navigation: Interpreting visually-grounded navigation instructions in real
  environments}}.
\newblock In \emph{CVPR 2018}.

\bibitem[{Bernardi et~al.(2016)Bernardi, Cakici, Elliott, Erdem, Erdem,
  Ikizler-Cinbis, Keller, Muscat, and Plank}]{Bernardietal:automatic}
Raffaella Bernardi, Ruket Cakici, Desmond Elliott, Aykut Erdem, Erkut Erdem,
  Nazli Ikizler-Cinbis, Frank Keller, Adrian Muscat, and Barbara Plank. 2016.
\newblock \href {http://dl.acm.org/citation.cfm?id=3013558.3013571} {Automatic
  description generation from images: A survey of models, datasets, and
  evaluation measures}.
\newblock \emph{J. Artif. Int. Res.}, 55(1):409--442.

\bibitem[{Bu{\ss} and Schlangen(2010)}]{bussetal:subutt}
Okko Bu{\ss} and David Schlangen. 2010.
\newblock Modelling sub-utterance phenomena in spoken dialogue systems.
\newblock In \emph{Proceedings of the 14th International Workshop on the
  Semantics and Pragmatics of Dialogue (Pozdial 2010)}, pages 33--41, Poznan,
  Poland.

\bibitem[{Byron et~al.(2007)Byron, Koller, Oberlander, Stoia, and
  Striegnitz}]{byron2007generating}
Donna Byron, Alexander Koller, Jon Oberlander, Laura Stoia, and Kristina
  Striegnitz. 2007.
\newblock Generating instructions in virtual environments (give): A challenge
  and an evaluation testbed for nlg.
\newblock \emph{Position Papers}, page~3.

\bibitem[{Chen and Lawrence~Zitnick(2015)}]{chen2015mind}
Xinlei Chen and C~Lawrence~Zitnick. 2015.
\newblock Mind's eye: A recurrent visual representation for image caption
  generation.
\newblock In \emph{Proceedings of the IEEE Conference on Computer Vision and
  Pattern Recognition}, pages 2422--2431.

\bibitem[{Clark(1996{\natexlab{a}})}]{clark:ul}
Herbert~H. Clark. 1996{\natexlab{a}}.
\newblock \emph{Using Language}.
\newblock Cambridge University Press, Cambridge.

\bibitem[{Clark(1996{\natexlab{b}})}]{clark1996using}
Herbert~H Clark. 1996{\natexlab{b}}.
\newblock Using language. 1996.
\newblock \emph{Cambridge University Press: Cambridge}, pages 274--296.

\bibitem[{Clark et~al.(1991)Clark, Brennan et~al.}]{clark1991grounding}
Herbert~H Clark, Susan~E Brennan, et~al. 1991.
\newblock Grounding in communication.
\newblock \emph{Perspectives on socially shared cognition}, 13(1991):127--149.

\bibitem[{Clark and Wilkes-Gibbs(1986)}]{clarkwilkes:ref}
Herbert~H. Clark and Deanna Wilkes-Gibbs. 1986.
\newblock Referring as a collaborative process.
\newblock \emph{Cognition}, 22:1--39.

\bibitem[{C{\^{o}}t{\'{e}} et~al.(2018)C{\^{o}}t{\'{e}}, K{\'{a}}d{\'{a}}r,
  Yuan, Kybartas, Barnes, Fine, Moore, Hausknecht, Asri, Adada, Tay, and
  Trischler}]{Cote2018}
Marc-Alexandre C{\^{o}}t{\'{e}}, {\'{A}}kos K{\'{a}}d{\'{a}}r, Xingdi Yuan, Ben
  Kybartas, Tavian Barnes, Emery Fine, James Moore, Matthew Hausknecht,
  Layla~El Asri, Mahmoud Adada, Wendy Tay, and Adam Trischler. 2018.
\newblock \href {https://doi.org/EC.00123-10 [pii]\r10.1128/EC.00123-10}
  {{TextWorld: A Learning Environment for Text-based Games}}.
\newblock \emph{ArXiv}.

\bibitem[{Das et~al.(2017{\natexlab{a}})Das, Datta, Gkioxari, Lee, Parikh, and
  Batra}]{dasetal:eqa}
Abhishek Das, Samyak Datta, Georgia Gkioxari, Stefan Lee, Devi Parikh, and
  Dhruv Batra. 2017{\natexlab{a}}.
\newblock \href {http://arxiv.org/abs/1711.11543} {Embodied question
  answering}.
\newblock \emph{CoRR}, abs/1711.11543.

\bibitem[{Das et~al.(2017{\natexlab{b}})Das, Kottur, Gupta, Singh, Yadav,
  Moura, Parikh, and Batra}]{das2017visual}
Abhishek Das, Satwik Kottur, Khushi Gupta, Avi Singh, Deshraj Yadav,
  Jos{\'e}~MF Moura, Devi Parikh, and Dhruv Batra. 2017{\natexlab{b}}.
\newblock Visual dialog.
\newblock In \emph{Proceedings of the IEEE Conference on Computer Vision and
  Pattern Recognition}, volume~2.

\bibitem[{De~Vries et~al.(2017)De~Vries, Strub, Chandar, Pietquin, Larochelle,
  and Courville}]{vries2017guesswhat}
Harm De~Vries, Florian Strub, Sarath Chandar, Olivier Pietquin, Hugo
  Larochelle, and Aaron Courville. 2017.
\newblock Guesswhat?! visual object discovery through multi-modal dialogue.
\newblock In \emph{Proc. of CVPR}.

\bibitem[{Devlin et~al.(2015)Devlin, Cheng, Fang, Gupta, Deng, He, Zweig, and
  Mitchell}]{devlin:imcaqui}
Jacob Devlin, Hao Cheng, Hao Fang, Saurabh Gupta, Li~Deng, Xiaodong He,
  Geoffrey Zweig, and Margaret Mitchell. 2015.
\newblock Language models for image captioning: The quirks and what works.
\newblock In \emph{Proceedings of the 53rd Annual Meeting of the Association
  for Computational Linguistics and the 7th International Joint Conference on
  Natural Language Processing (Volume 2: Short Papers)}, pages 100--105,
  Beijing, China. Association for Computational Linguistics.

\bibitem[{Fang et~al.(2015)Fang, Gupta, Iandola, Srivastava, Deng, Dollar, Gao,
  He, Mitchell, Platt, Zitnick, and Zweig}]{fangetal:2015}
Hao Fang, Saurabh Gupta, Forrest Iandola, Rupesh Srivastava, Li~Deng, Piotr
  Dollar, Jianfeng Gao, Xiaodong He, Margaret Mitchell, John Platt, Lawrence
  Zitnick, and Geoffrey Zweig. 2015.
\newblock From captions to visual concepts and back.
\newblock In \emph{Proceedings of CVPR}, Boston, MA, USA. IEEE.

\bibitem[{Fern\'andez and Schlangen(2007)}]{fernangen:sigd07}
Raquel Fern\'andez and David Schlangen. 2007.
\newblock Referring under restricted interactivity conditions.
\newblock In \emph{Proceedings of the 8th SIGdial Workshop on Discourse and
  Dialogue}, pages 136--139, Antwerp, Belgium.

\bibitem[{Haber et~al.(2019)Haber, Baumg{\"{a}}rtner, Takmaz, Gelderloos,
  Bruni, and Fern{\'{a}}ndez}]{Haber2019}
Janosch Haber, Tim Baumg{\"{a}}rtner, Ece Takmaz, Lieke Gelderloos, Elia Bruni,
  and Raquel Fern{\'{a}}ndez. 2019.
\newblock {The PhotoBook Dataset: Building Common Ground through
  Visually-Grounded Dialogue}.
\newblock In \emph{Proceedings of the 2019 meeting of the Association for
  Computational Linguistics}, Florence, Italy.

\bibitem[{Harnard(1990)}]{harnard:grounding}
Stevan Harnard. 1990.
\newblock The symbol grounding problem.
\newblock \emph{Physica D}, 42:335--346.

\bibitem[{Kazemzadeh et~al.(2014)Kazemzadeh, Ordonez, Matten, and
  Berg}]{Kazemzadeh2014}
Sahar Kazemzadeh, Vicente Ordonez, Mark Matten, and Tamara~L Berg. 2014.
\newblock {ReferItGame: Referring to Objects in Photographs of Natural Scenes}.
\newblock In \emph{Proceedings of the Conference on Empirical Methods in
  Natural Language Processing (EMNLP 2014)}, pages 787--798, Doha, Qatar.

\bibitem[{Kiela and Bottou(2014)}]{kiela:2014}
Douwe Kiela and L\'{e}on Bottou. 2014.
\newblock \href {http://www.aclweb.org/anthology/D14-1005} {Learning image
  embeddings using convolutional neural networks for improved multi-modal
  semantics}.
\newblock In \emph{Proceedings of the 2014 Conference on Empirical Methods in
  Natural Language Processing (EMNLP)}, pages 36--45, Doha, Qatar. Association
  for Computational Linguistics.

\bibitem[{Kirsh and Maglio(1994)}]{Kirsh1994}
David Kirsh and Paul Maglio. 1994.
\newblock \href {https://doi.org/10.1207/s15516709cog1804_1} {{On
  Distinguishing Epistemic from Pragmatic Action}}.
\newblock \emph{Cognitive Science}, 18(4):513--549.

\bibitem[{Lazaridou et~al.(2015)Lazaridou, Pham, and
  Baroni}]{lazaridou-pham-baroni:2015}
Angeliki Lazaridou, Nghia~The Pham, and Marco Baroni. 2015.
\newblock \href {http://www.aclweb.org/anthology/N15-1016} {Combining language
  and vision with a multimodal skip-gram model}.
\newblock In \emph{Proceedings of the 2015 Conference of the North American
  Chapter of the Association for Computational Linguistics: Human Language
  Technologies}, pages 153--163, Denver, Colorado. Association for
  Computational Linguistics.

\bibitem[{Ma et~al.(2019)Ma, Lu, Wu, AlRegib, Kira, Socher, and Xiong}]{Ma2019}
Chih-Yao Ma, Jiasen Lu, Zuxuan Wu, Ghassan AlRegib, Zsolt Kira, Richard Socher,
  and Caiming Xiong. 2019.
\newblock \href {http://arxiv.org/abs/1901.03035} {{Self-Monitoring Navigation
  Agent via Auxiliary Progress Estimation}}.
\newblock \emph{ArXiv}, pages 1--18.

\bibitem[{Mao et~al.(2015)Mao, Huang, Toshev, Camburu, Yuille, and
  Murphy}]{mao15}
Junhua Mao, Jonathan Huang, Alexander Toshev, Oana Camburu, Alan~L. Yuille, and
  Kevin Murphy. 2015.
\newblock \href {http://arxiv.org/abs/1511.02283} {Generation and comprehension
  of unambiguous object descriptions}.
\newblock \emph{CoRR}, abs/1511.02283.

\bibitem[{Rosenberg and Cohen(1964)}]{Rosenberg1964}
Seymour Rosenberg and Bertram~D. Cohen. 1964.
\newblock {Speakers' and Listeners' Processes in a Word-Communication Task}.
\newblock \emph{Science}, 145(3637):1201--1204.

\bibitem[{Savva et~al.(2017)Savva, Chang, Dosovitskiy, Funkhouser, and
  Koltun}]{savva2017minos}
Manolis Savva, Angel~X. Chang, Alexey Dosovitskiy, Thomas Funkhouser, and
  Vladlen Koltun. 2017.
\newblock {MINOS}: Multimodal indoor simulator for navigation in complex
  environments.
\newblock \emph{arXiv:1712.03931}.

\bibitem[{Savva et~al.(2019)Savva, Kadian, Maksymets, Zhao, Wijmans, Jain,
  Straub, Liu, Koltun, Malik, Parikh, and Batra}]{habitat19arxiv}
Manolis Savva, Abhishek Kadian, Oleksandr Maksymets, Yili Zhao, Erik Wijmans,
  Bhavana Jain, Julian Straub, Jia Liu, Vladlen Koltun, Jitendra Malik, Devi
  Parikh, and Dhruv Batra. 2019.
\newblock \href {http://arxiv.org/abs/1904.01201} {{Habitat: A Platform for
  Embodied AI Research}}.
\newblock \emph{ArXiv}.

\bibitem[{Schlangen(2016)}]{schlangen:justification}
David Schlangen. 2016.
\newblock {Grounding, Justification, Adaptation: Towards Machines That Mean
  What They Say}.
\newblock In \emph{Proceedings of the 20th Workshop on the Semantics and
  Pragmatics of Dialogue (JerSem)}.

\bibitem[{Schlangen et~al.(2018)Schlangen, Diekmann, Ilinykh, and
  Zarrieß}]{slurk.semdial18}
David Schlangen, Tim Diekmann, Nikolai Ilinykh, and Sina Zarrieß. 2018.
\newblock {slurk – A Lightweight Interaction Server For Dialogue Experiments
  and Data Collection}.
\newblock In \emph{Short Paper Proceedings of the 22nd Workshop on the
  Semantics and Pragmatics of Dialogue (AixDial / semdial 2018)}.

\bibitem[{Schlangen et~al.(2016)Schlangen, Zarriess, and
  Kennington}]{schlazar:acl16}
David Schlangen, Sina Zarriess, and Casey Kennington. 2016.
\newblock Resolving references to objects in photographs using the
  words-as-classifiers model.
\newblock In \emph{Proceedings of the 54rd Annual Meeting of the Association
  for Computational Linguistics (ACL 2016)}.

\bibitem[{Silberer and Lapata(2014)}]{silberer-lapata:2014}
Carina Silberer and Mirella Lapata. 2014.
\newblock \href {http://www.aclweb.org/anthology/P14-1068} {Learning grounded
  meaning representations with autoencoders}.
\newblock In \emph{Proceedings of the 52nd Annual Meeting of the Association
  for Computational Linguistics (Volume 1: Long Papers)}, pages 721--732,
  Baltimore, Maryland. Association for Computational Linguistics.

\bibitem[{Stent(2011)}]{stent2011computational}
Amanda~J Stent. 2011.
\newblock Computational approaches to the production of referring expressions:
  Dialog changes (almost) everything.
\newblock In \emph{PRE-CogSci Workshop}.

\bibitem[{Sutton and Barto(1998)}]{suttonbarto:rl}
Richard~S. Sutton and Andrew~G. Barto. 1998.
\newblock \emph{Reinforcement Learning}.
\newblock MIT Press, Cambridge, USA.

\bibitem[{Szegedy et~al.(2015)Szegedy, Liu, Jia, Sermanet, Reed, Anguelov,
  Erhan, Vanhoucke, and Rabinovich}]{googlenet}
Christian Szegedy, Wei Liu, Yangqing Jia, Pierre Sermanet, Scott Reed, Dragomir
  Anguelov, Dumitru Erhan, Vincent Vanhoucke, and Andrew Rabinovich. 2015.
\newblock Going deeper with convolutions.
\newblock In \emph{CVPR 2015}, Boston, MA, USA.

\bibitem[{Tokunaga et~al.(2012)Tokunaga, Iida, Terai, and
  Kuriyama}]{tokunaga2012rex}
Takenobu Tokunaga, Ryu Iida, Asuka Terai, and Naoko Kuriyama. 2012.
\newblock The rex corpora: A collection of multimodal corpora of referring
  expressions in collaborative problem solving dialogues.
\newblock In \emph{Proceedings of the Eighth International Conference on
  Language Resources and Evaluation (LREC 2012)}.

\bibitem[{Urbanek et~al.(2019)Urbanek, Fan, Karamcheti, Jain, Humeau, Dinan,
  Rockt{\"{a}}schel, Kiela, Szlam, and Weston}]{Urbanek2019}
Jack Urbanek, Angela Fan, Siddharth Karamcheti, Saachi Jain, Samuel Humeau,
  Emily Dinan, Tim Rockt{\"{a}}schel, Douwe Kiela, Arthur Szlam, and Jason
  Weston. 2019.
\newblock \href {http://arxiv.org/abs/1903.03094} {{Learning to Speak and Act
  in a Fantasy Text Adventure Game}}.
\newblock \emph{ArXiv}.

\bibitem[{Vinyals et~al.(2015)Vinyals, Toshev, Bengio, and
  Erhan}]{vinyals:show}
Oriol Vinyals, Alexander Toshev, Samy Bengio, and Dumitru Erhan. 2015.
\newblock Show and tell: A neural image caption generator.
\newblock In \emph{Computer Vision and Pattern Recognition}.

\bibitem[{Williams and Young(2007)}]{williamsyoung:POMDPs}
Jason Williams and Steve Young. 2007.
\newblock Partially observable {M}arkov decision processes for spoken dialog
  systems.
\newblock \emph{Computer Speech and Language}, 21(2):231--422.

\bibitem[{Yu et~al.(2016)Yu, Poirson, Yang, Berg, and Berg}]{Yu2016}
Licheng Yu, Patrick Poirson, Shan Yang, Alexander~C. Berg, and Tamara~L. Berg.
  2016.
\newblock \href {https://doi.org/10.1007/978-3-319-46475-6_5} {\emph{Modeling
  Context in Referring Expressions}}, pages 69--85. Springer International
  Publishing, Cham.

\bibitem[{Zarrie{\ss} et~al.(2016)Zarrie{\ss}, Hough, Kennington,
  Manuvinakurike, DeVault, Fernandez, and Schlangen}]{zarriess2016pentoref}
Sina Zarrie{\ss}, Julian Hough, Casey Kennington, Ramesh Manuvinakurike, David
  DeVault, Raquel Fernandez, and David Schlangen. 2016.
\newblock Pentoref: A corpus of spoken references in task-oriented dialogues.
\newblock In \emph{10th edition of the Language Resources and Evaluation
  Conference}.

\bibitem[{Zhou et~al.(2017)Zhou, Zhao, Puig, Fidler, Barriuso, and
  Torralba}]{zhou2017scene}
Bolei Zhou, Hang Zhao, Xavier Puig, Sanja Fidler, Adela Barriuso, and Antonio
  Torralba. 2017.
\newblock Scene parsing through ade20k dataset.
\newblock In \emph{Proceedings of the IEEE Conference on Computer Vision and
  Pattern Recognition}.

\end{thebibliography}
}

\end{document}